\documentclass[letterpaper]{article}
\usepackage{iccc}

\usepackage{times}
\usepackage{helvet}
\usepackage{courier}
\usepackage{graphicx}

\usepackage{multirow}
\usepackage{makecell}

\title{Evaluating Creativity in Computational Co-Creative Systems}
\author{Pegah Karimi$^1$, Kazjon Grace$^2$, Mary Lou Maher$^1$, Nicholas Davis$^1$\\
$^1$UNC Charlotte, $^2$The University of Sydney\\
$^1$USA, $^2$Australia\\
pkarimi@uncc.edu, kazjon.grace@sydney.edu.au, m.maher@uncc.edu, ndavis64@uncc.edu\\
}

\begin{document} 
\maketitle
\begin{abstract}
\begin{quote}
This paper provides a framework for evaluating creativity in co-creative systems: those that involve computer programs collaborating with human users on creative tasks. We situate co-creative systems within a broader context of computational creativity and explain the unique qualities of these systems. We present four main questions that can guide evaluation in co-creative systems: Who is evaluating the creativity, what is being evaluated, when does evaluation occur and how the evaluation is performed. These questions provide a framework for comparing how existing co-creative systems evaluate creativity, and we apply them to examples of co-creative systems in art, humor, games and robotics. We conclude that existing co-creative systems tend to focus on evaluating the user experience. Adopting evaluation methods from autonomous creative systems may lead to co-creative systems that are self-aware and intentional.  
\end{quote}
\end{abstract}

\section{Introduction}

Creative systems are intelligent systems that can perform creative tasks alone or in collaboration. These systems can enable a wide variety of tasks with a similarly wide variety of roles for human participants. There are three main strategies by which the role of humans in creative systems can be characterized:  fully autonomous systems, creativity support tools, and co-creative systems. 

Fully autonomous systems are built to generate creative artifacts that are judged by users to be creative \cite{elgammal2017can,colton2015painting}. These systems are based on a variety of technologies, from corpus-trained statistical machine learning techniques, to production rules, to evolutionary approaches or planning based systems, all designed to produce output that is judged as creative by some evaluation process.

\begin{figure}[ht!]
\centering
\includegraphics[width=0.48\textwidth]{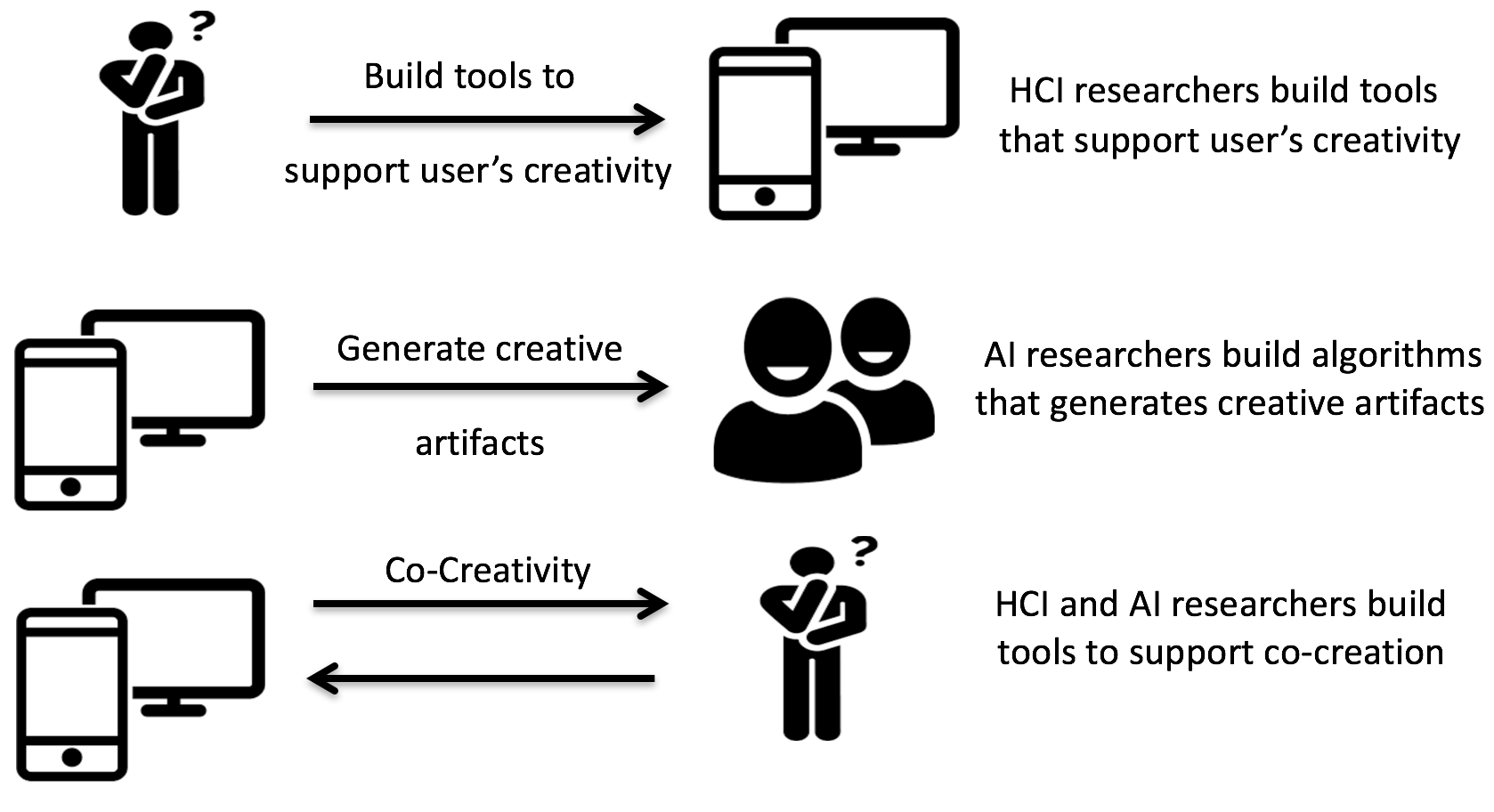}
\caption{Three main trends in creative systems: creativity support tools, fully autonomous systems and co-creative systems.}
\end{figure}

Creativity support tools, on the other hand, are tools and apps that are built in order to support the user's creativity \cite{compton2015casual,hoffman2010gesture}. Shneiderman \shortcite{shneiderman2007creativity} defines creativity support tools as tools that develop the creative thought of users and allow them to be both productive and innovative. In his work, he has introduced a set of design principles specifically for supporting user's creativity. Some of these principles include supporting simplicity, wide range of exploration, and different paths and styles. There is no requirement in this definition that these tools be pro-active in the creative process, much less aware of the creativity or quality of their own output. Arguably, the interpretation of the above definition says that a paintbrush meets the requirement of a creativity support tool.

Co-creativity is when computers and humans collaborate with each other to build shared creative artifacts \cite{wen2015omg,davis2015enactive}. The term evolved from referring to any collaborative creative activity to referring purely to those involving at least one computational actor, and can be considered a contraction of ``computational co-creativity''. It involves different types of collaboration (e.g. division of labor, assistantship, partnership) between multiple parties where at least one of the parties is an AI agent. In these systems, each agent has to perceive other agents' contributions and express its own creative ideas through autonomous action. In this research we define a co-creative system as: \textit{Interaction between at least one AI agent and at least one human where they take action based  on the response of their partner and their own conceptualization of creativity during the co-creative task.}

There are various applications of co-creativity in domains including arts (Jacob et al, 2013), games (Lucas and Martinho, 2017), robotics (Hoffman and Weinberg, 2010) and humor (Wen et al, 2015). While in most of the literature the focus is on the design and implementation of these systems, there is less research investigating how these systems can be evaluated. In this paper we characterize the different ways that co-creative systems can be evaluated, aiming to give clarity to current and future research in this rapidly evolving field.

We present four main questions to compare the evaluation of co-creative systems. The first question focuses on \textbf{who} evaluates the creativity, e.g. the system itself, human judges, etc.. The second question focuses on \textbf{what} is being evaluated, such as the creative interaction and the creative artifact. The third question focuses on \textbf{when} the evaluation is done: is it formative or summative? The last question focuses on \textbf{how} the evaluation is performed, e.g. methods and metrics. 

This paper is organized as follows: The first section describes co-creative systems. The second section focuses on the design and implementation of co-creative systems in different domains. The third section discusses the evaluation of co-creative systems and finally the last section addresses how the evaluation is done in each of the applications that were discussed in section two. The main contribution of this work is the articulation of a framework for evaluating co-creative systems. We also identify a need for co-creative systems to adopt methods and metrics for evaluating the creativity of creative agents to distinguish co-creativity from creativity support.

\section{Co-Creative Systems}

Co-creative systems are one of the growing trends in creative AI, in which computers and users interact with each other to make creative artifacts. Co-creativity is a type of collaboration where the contributions from different parties are synthesized and added upon during the interaction. Some forms of collaboration, such as division of labor, involve individuals working independently and sharing their ideas after accomplishing tasks. In the majority of co-creative systems to date, the collaboration between participants is done in real time during the task. \citeauthor{davis2015enactive} \shortcite{davis2015enactive} establishes synchronous collaboration as a requirement, defining co-creativity as a process where users and computers can collaboratively improvise on a shared artifact \textit{during the creative process}. 

Another similar term is called mixed initiative co-creativity \cite{yannakakis2014mixed}. In his definition, both the human and the computer take initiative in creating a new artifact, meaning both parties are actively contributing to the shared artifact. ``Actively contributing'' in a mixed-initiative system means that the computational agent(s) contribute proactively, rather than solely in response to a user request.  The human and artificial agents do not need to contribute to the same degree and there is no need for their contribution to be symmetrical.  


Mixed-initiative systems are by definition co-creative, but not all co-creative systems are mixed-initiative. In many systems there is an explicit turn-taking process, but this is not a requirement: some systems are machine-initiative dominated, operating as a kind of ``wizard'' interface in which the user is consulted during a highly scripted process, while others are user-dominated, with the system jumping in only infrequently with suggestions or critique.


\section{Examples of Co-Creative Systems}

Co-creativity has been applied in domains as broad as art, humor, game and robotics. Two examples of such systems are the Drawing Apprentice \cite{davis2015enactive} and ViewPoints AI \cite{jacob2015viewpoints}. The Drawing Apprentice is a co-creative drawing application in which there is a collaboration between the user and an AI agent on a drawing task. In this system, the user starts drawing a sketch on the canvas and the agent responds by adding to the user's input in real time. ViewPoints AI is an artistic co-creative system for the performing arts.  The user starts dancing and the system projects a life-sized silhouette that dances back, both following the user's cues and initiating its own. 

Examples of co-creative systems in games include the Sentient Sketchbook \cite{yannakakis2014mixed} and 3Buddy \cite{lucas2017stay}. Sentient Sketchbook is a mixed-initiative game level design tool that fosters user creativity. Human designers can create game levels, and the AI agent responds in real time with suggested additions and modifications. 3Buddy assists its human user in generating game levels, following three different goals to do so: 1) converging towards the user's emerging design, 2) innovating on that design, and 3) working within the guidelines explicitly stated by the user. 

Cahoots is a co-creative humor system \cite{wen2015omg}. It operates as a web-based chat platform in which two users and an AI agent collaborate through a conversation to foster humor. The users send text messages to each other, including humorous in-line images if they desire, and the AI interjects with additional images. 

Shimon is a co-creative robot in the domain of music \cite{hoffman2010gesture}.  Its authors describe it as an interactive improvisational robotic musician. The robot listens and responds to a musician in real time. 

\section{Evaluating Computational Co-Creativity}

Evaluating computational models of creativity is an important component of designing and understanding creative systems \cite{jordanous2012standardised}. Evaluating co-creative systems is still an open research question and there is no standard metric that can be used across specific systems. Below we present 4 questions that can serve to characterize the many and varied approaches to evaluating computational models of co-creativity. 

\subsection{Who is evaluating the creativity?}

When asking who evaluates the creativity in a co-creative system there are three broad categories of answer: the AI, the user and a third party. We refer to the AI evaluating output as self-evaluation: it is aware of its own creativity during the creative process. This represents a kind of metacognition \cite{cox2011metareasoning}, or thinking about thinking: the system is aware its own processes, and can be considered to be intentional \cite{colton2008creativity}. 

\citeauthor{grace2016surprise} \shortcite{grace2016surprise} introduce an evaluation method called surprise-triggered reformulation, in which this metacognitive self-evaluation triggers the formation of new design goals. \citeauthor{karimi2018deep} \shortcite{karimi2018deep} proposes a method for identifying and introducing conceptual shifts in a co-creative drawing context. These systems demonstrate the potential for co-creativity with self-evaluation.

Situating the focus of evaluation in co-creativity within the user can introduce a new set of affordances for interaction during creative tasks. In this approach users judge the creativity of the system or its outputs. In ViewPoints AI, a user study is conducted after the interaction to determine the user's level of engagement, an offline approach to user evaluation \cite{jacob2015viewpoints}. As an example of evaluation during the creative task, in the Drawing Apprentice  the user votes (like, dislike) on sketches as they are generated by the agent \cite{davis2015enactive}.  

\begin{figure*}[ht!]
\centering
\includegraphics[width=1\textwidth,height=2.2in]{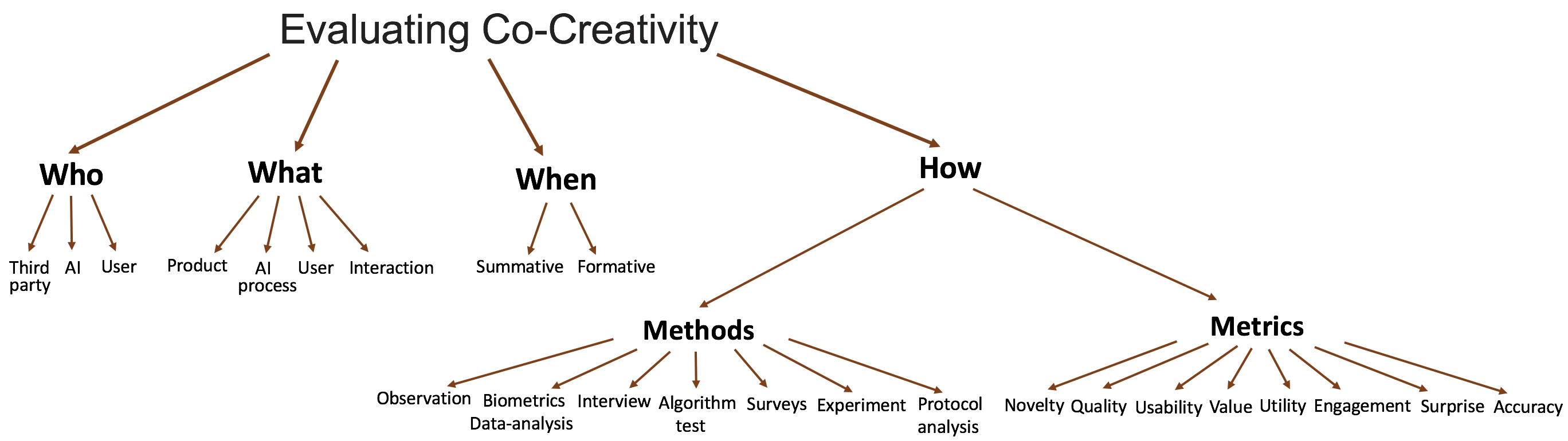}
\caption{A hierarchical tree of evaluating creativity in computational co-creative systems.}
\end{figure*}

The last category, third-party, is when evaluation is judged by neither the system nor its user. This kind of evaluation often takes the form of domain experts evaluating the quality or creativity of the result or product.  This is particularly useful in domains where substantial knowledge or expertise is required to effectively judge creative artifacts. \citeauthor{yannakakis2014mixed} \shortcite{yannakakis2014mixed} performed a user study of this kind of a co-creative game level design tool by asking experts to judge the creativity of the resultant levels.  Another approach to third-party design is devolving the evaluative responsibility to the users of the output (as distinct from the users of the system, the co-creators). 


\subsection{What is being evaluated?} Evaluations of co-creative systems can, like other creative systems, focus on the evaluation of the product, the process, and user creativity, but they can also focus on evaluating the interactions between the user and the system. Broadly, evaluations of process, product, and user creativity are similar enough in co-creative contexts to benefit from the rich history of research in autonomous creative systems \cite{grace2015data,jordanous2012standardised,saunders2001artificial,schmidhuber2008driven,wiggins2006preliminary} and studies of human creativity \cite{besemer1999confirming,cropley2005engineering}.  We discuss here specific issues relevant to co-creativity.

The artifact(s) resulting from the collaboration represent the combined effort of the user and system, which we refer to as the ``product''. In more goal-directed creative tasks, the user and system are both working towards a common goal. However, in more open-ended creative tasks, the user and system can improvise on shared or independent goals in an exploration where emergent creativity can occur. A game level design tool like 3Buddy is an example of goal-directed product-based evaluation.  A collaborative sketch tool like the Drawing Apprentice is an example of the latter kind: evaluation of the artifacts that result from a more open-ended exploratory creative task. 

Evaluating the creativity of the system or the process refers to the software or the computational model that has been built for a particular system. \citeauthor{colton2008creativity} \shortcite{colton2008creativity} evaluates the creativity of the software based on skill, appreciation and imagination. Skill refers to the ability of the software to create products, it captures traditional notions of the ``craft'' embodied in a particular creative domain. Appreciation indicates the ability of the software to detecting particular patterns in generated artifacts: its ability to self-evaluate. Lastly, imagination refers to the ability of the software to construct a new specific representation from existing artifacts. Evaluating the creative process used within an autonomous creative system is a challenging prospect, as is evaluating the creative processes of a human, although for very different reasons.  Evaluating the creativity of the processes used in a co-creative system combines the difficulties of both. 

A critical component of co-creative systems is the interaction between machine and human.  Evaluating this interface (for usability, expressiveness, effectiveness, or the affect it produces) is the final focus of what can be evaluated in co-creative systems. There is a dynamism to the interaction between user and system that is an innate part of all creative collaborations. Evaluating these interactions requires a very different set of methods to evaluating either the creative product or the creative process. \citeauthor{davis2017creative} \shortcite{davis2017creative} introduced ``creative sense-making'', a cognitive model of the interaction dynamics between the user and the agent during a drawing task. User behavior was evaluated as either ``clamped'' (in direct engagement with the creative artifact), or ``unclamped'' (not directly controlling the artifact -- observing, reflecting, or disengaged).  This representation of sequences of types of engagement begins to characterize the process co-creative interaction.

\subsection{When does evaluation occur?} Creativity evaluation can be formative (i.e. performed during the process) or summative (i.e. performed after the creative process). In autonomous creative systems formative evaluation is typically part of a generate-and-test loop, providing the system with the feedback that guides its search.  In co-creativity the possibilities of formative evaluation are substantially broader, given that now the emerging proto-artifacts are shared between human and AI. The user can evaluate its own output or that of the user, and vice versa. This can be used as a way for each participant to attempt to guide the other, and can occur in a wide variety of turn-taking, real-time, mixed-initiative and other contexts. For example, in 3Buddy \cite{lucas2017stay}, users judge the creativity of the level design at each step of a generative system that uses an evolutionary algorithm, providing input from the user to accompany the system's automated formative self-evaluation.

Summative evaluation of creativity plays a very different role.  In some contexts the user provides feedback that might influence future tasks.  In others an evaluation is performed as part of the experimental context surrounding a system. Summative analyses of creativity performed in this latter context blur the line between being evaluation and being validation: are they part of the system, or part of the research, or both?  


\subsection{How is evaluation performed?} 
\subsubsection{Methods} The primary method of evaluation in the co-creativity literature to date has been user studies.  The specific approach to performing those user studies has been quite varied, including protocol analysis, survey data, interview, experiment and observation.   

Protocol analysis is an empirical method in which the user's behavior with the system is characterized and analyzed. Protocol analysis is used in design science and design cognition research, in which coding schemes are applied to segment and categorize the sequence of physical, digital and verbal actions that comprise creative tasks.  Originally a ``design protocol'', whether concurrent or retrospective, was a transcript of the think aloud method in which the designer was asked to talk while designing (concurrent) or while viewing a recording of their design task (retrospective). More recently, a design protocol is associated with any sequential recording of a design task, including speech, gesture, body movement, facial expressions, dialogue and digital actions. When evaluating the Drawing Apprentice where users were asked to collaborate on drawing tasks, users were asked to view a recording of themselves designing and describe their thought processes for each action, which were then categorized according to the coding scheme \cite{davis2015enactive}.  This video walkthrough is an example of a retrospective protocol.

Surveys are a method of obtaining data from the users of co-creative system that are much more scalable but less rich than protocol analysis.  Surveys can take different forms, but their common goal is to obtain insight into user perceptions of the creative system and the creative tasks. This can include system usability, self-reflection, evaluation of the output, and evaluation of the system's processes. An example of survey data can be found in Sentient Sketchbook \cite{yannakakis2014mixed}, where users were asked about the usability of the game level design tool. In that study users were generally positive about the tool's interface and their interactions with the co-creative system. 

Interviews are a qualitative method for evaluating user perceptions of co-creative systems, providing an interpretive alternative to the quantitative and empirical protocol analysis methods. More specifically, these are typically semi-structured interviews, a method common in the social sciences and in human-computer interaction research as a way to elicit rich and nuanced perceptions from small groups of users. In ViewPoints AI \cite{jacob2015viewpoints}, interviews showed that users expected the agent to respond to each of their movements in real-time and were disappointed on the occasions where it did not. 


Observational methods are another common evaluation method. Observing creative tasks without intervening or pre-committing to a specific coding scheme enables investigation of a broad range of behaviors. Examples from co-creative systems include Shimon \cite{hoffman2010gesture}, where observation showed that the source of inspiration for the current moment of performance alternated between the human and the robot player.

The last user study method that has been successfully employed in the study of creativity support tools, but not yet applied to co-creative systems, is the use of biometric data to quantify human creativity. For example, \citeauthor{carroll2012triangulating} \shortcite{carroll2012triangulating} utilized electroencephalogram (EEG) to measure neural signals during a creative task. This work sought to measure 'in-the-moment creativity' (ITMC), which is defined as periods of heightened creativity during the creative process. The EEG data was combined with self-report data about the user's creative state to triangulate when users where experiencing moments of high creativity. This study demonstrates the potential for biometric data to be applied to co-creative systems to help quantify user creativity while interacting with the system. 

In addition to user studies, researchers have also tested the algorithms themselves to determine their efficacy. This testing process validates the algorithms and models used by the AI agent employed in the co-creative system. For example, \citeauthor{singh2017unified} \shortcite{singh2017unified} performs a validation test on the object recognition and generation algorithms used in a co-creative drawing application. This type of validation is common in the machine learning literature to test the effectiveness of the algorithm. In a co-creative context, this information can be used to tweak the algorithm to better suit the needs of the co-creative system. 

\subsubsection{Metrics} The set of  metrics for developing computational models for evaluating creativity is very broad, including those defined in \cite{lamb2018evaluating,grace2015data,ritchie2007some,wiggins2006preliminary}. In response to a focus on novelty and value as the hallmark of creativity that started as early as \cite{newell1959processes}, \citeauthor{maher2012using} \shortcite{maher2012using} add a third dimension called surprise, which quantifies how unexpected the creative product is given the sequence of decisions or products that have recently occurred.  


\citeauthor{pease2011impact} \shortcite{pease2011impact} introduce two different levels for evaluation: cultural value of the outcome (a measure of product) and the complexity of the system's behavior (a measure of process). \cite{francca2016regent} argue that evaluating computational creativity should be domain independent. They introduces a metric, called Regent-Dependent Creativity (RDC), in which generated artifacts are represented as dependency pairs. RDC  measures novelty and value within this structure.

In more recent literature, researchers aim at operationalizing creativity  by building computational models. \citeauthor{agres2015conceptualizing} \shortcite{agres2015conceptualizing} introduces a computational linguistic model that maps word representations into a conceptual space. The model is based on word co-occurrence in the context of music and poetry. In order to validate the accuracy of the model, user responses to word association is also recorded and compared with the computational model results. \citeauthor{grace2015data} \shortcite{grace2015data} introduces a probabilistic model in order to compute the surprise value in the domain of mobile devices. The model captures the degree of unexpectedness of the observed artifact. These models imply that the less likely an event or combination of events occurs the more likely it is to be surprising. 

One important metric that makes co-creative systems different from other computational creativity systems is the engagement of the user with the system. In Viewpoints AI \cite{jacob2015viewpoints} the engagement of users was evaluated qualitatively, and was found to be highly positive. In  the Sentient Sketchbook \cite{yannakakis2014mixed} two metrics are developed: perceived usefulness and perceived quality. In 3Buddy \cite{lucas2017stay} metrics for utility and efficiency are developed. Many co-creative systems also measure usability, including ViewPoints AI \cite{jacob2015viewpoints} and the Sentient Sketchbook \cite{yannakakis2014mixed}.

One final family of metrics applied in co-creative metrics are those derived from accuracy, or more specifically the degree to which generated output matches a reference dataset.  These measures often originate from machine learning, where accuracy is a central concern. An example of this is from a recent extension of the Drawing Apprentice system \cite{singh2017unified}, in which the classification accuracy and generation loss of their model is reported on two different public datasets. 

\section{Case Studies of Co-creative Evaluation}

In this section we focus on how the evaluation is performed in different co-creative systems. Table 1 summarizes the above questions for six example systems.

\begin{table*}
\begin{center}
\footnotesize
\begin{tabular}{ | c | c | c | c | c | c |}
\hline
\thead{System} & \thead{Who} & \thead{When} & \thead{How \\ (Metric)} & \thead{How \\ (Method)} & \thead{What}\\
\hline
\multirow{2}{*}[-1.4em]{Drawing Apprentice} & \makecell{AI} & Summative & \makecell{Classification Accuracy \\ \& Generation Loss} & Algorithm testing  & \multirow{ 2}{*}[-1.1em]{\makecell{Product \& \\ Interactive experience}} \\
\cline{2-5}
 & Users & Formative & Usability &  \makecell{Voting (like, dislike), \\ Survey data \& \\ Retrospective protocol analysis}  &  \\
\hline
ViewPoints AI & Users & \makecell{Formative \& \\ Summative} & \makecell{Engagement \\ \& Usability}  &  \makecell{Observation}  & \makecell{Product \& \\ Interactive experience}\\
\hline
Sentient Sketchbook & Experts & \makecell{Formative \& \\ Summative} & \makecell{Usefulness, Quality \\ \& Usability}  &  \makecell{Protocol Analysis, \\ Survey Data, Interview\\ Experiment and\\  Observation}  & Product\\
\hline
3Buddy & \makecell{Users \& \\ Experts} & \makecell{Summative} & \makecell{Utility \& Efficiency}  &  \makecell{Survey Data, Interview\\ Experiment \&\\  Observation}  & Product\\
\hline
CAHOOTS & \makecell{Users} & \makecell{Summative} & \makecell{Usability}  &  \makecell{Survey Data \& \\ Experiment}  & Product\\
\hline
SHIMON & \makecell{Users} & \makecell{Summative} & \makecell{Engagement}  &  \makecell{Observation}  & Product\\
\hline
\end{tabular}
\caption{Answers to questions in section three for six different co-creative systems. Note that two studies involving the Drawing Apprentice were published, using different evaluation methods.}

\end{center}
     
\end{table*}

\subsection{Evaluating creativity in the Drawing Apprentice} In the Drawing Apprentice, several evaluation methods have been deployed, including both formative and summative user studies. Participants are first introduced to the unique features of the Drawing Apprentice system. As an example of formative evaluation, users are asked to rate sketches generated by the agent (like or dislike). This voting occurs at iterative steps when the agent responds to the user's input during the task.  For a summative evaluation of co-creativity, a combination of retrospective protocol analysis, interviews, and surveys were performed. Participants were asked to work with drawing apprentice for 12 minutes in two different sessions. In one session, they interact with the actual system and in the other they interact with a ``Wizard of Oz'' substitute (i.e. a fake system with a hidden human controller). After the task was complete, participants watched a video recording of their interaction and described what they were thinking at each point in the video during a retrospective protocol analysis. Then, participants were asked about their experiences through both interviews and surveys. The results show that the agent is able to coordinate with the user up to a certain degree as well as contributing to the user's drawing. 

In more recent work, a machine learning model called an Auxiliary Classifier Variational AutoEncoder (AC-VAE) was added to the co-creative system that allows the agent to classify and generate input images simultaneously in real time \cite{singh2017unified}. In this work, the evaluation is done offline through two metrics: classification accuracy and generation loss.  Both can be considered measures of value: the degree to which the system is able to categorize sketches made by the user, and the degree to which it is able to produce sketches that are similar to the user's sketch. Results are reported on two different public datasets in order to compare the accuracy of the AC-VAE model to other existing models. Their integration into the co-creative system and their impacts on user behavior and perception of creativity are still under development.

The formative evaluation of Drawing Apprentice leverages the voting system used by the machine learning algorithm in the system. This approach is interesting because it provides a method of evaluating how the agent is performing throughout the session without interrupting the creative flow of the user. It is possible to count how many times users clicked like/dislike, but this method is also unreliable as users do not have to use the voting at all. To get a more holistic understanding of the user's creative experience, the authors employed a retrospective protocol where participants watched their creative process and explained their thoughts. These videos can then be coded to understand themes and trends in the interaction. When supplemented with interviews and surveys, this type of user study can sketch an accurate description of the user's experience with the system. However, this analysis did not include a summative evaluation of the creative output of the interaction, which would help evaluate the relative creativity of both user and system. 

\subsection{Evaluating creativity in ViewPoints AI} 
The evaluation of this system is done by the users in a public space through a summative and formative user study. Participants are first presented with the prototype of the system and introduced to the features of the system through a demonstration of how to interact with the system. They are then asked to interact with the system without the aid of the researchers. During the interaction, researchers observed how participants interacted with the system (formative evaluation). After the interaction, participants provided feedback about their experiences (summative evaluation). The results show that users gave positive comments in terms of both the concept and the visual aesthetic of the system. The task observations show that the engagement of the users with the system was highly positive. However, participants were not always able to understand the intentions of the AI agent, with some participants not even understanding that the system was co-creative at all.  This highlights the need for AI agents to produce responses that are both similar and different enough to the user's movement. Another finding was that users expect immediate responses during turn-based interaction.

\subsection{Evaluating creativity in the Sentient Sketchbook} The evaluation of this system is done by the experts through formative and summative user studies. During this study the usability of the sentient sketchbook, game level design tool, is assessed. The evaluation is done online by sending the participants an email and receiving feedback via email as well. The study recruited five users to perform 24 different design sessions. Overall, feedback about the usability of the system were positive. 

For the summative evaluation of the creativity of the system, the evaluation is based on two metrics: degree of usefulness of the co-creative tool and quality of their interaction during the process. The first metric, degree of usefulness, refers to usability of design suggestions in different sessions. Based on the user feedback there were cases where the design suggestions were not useful. Particularly most design suggestions were selected in the beginning of the co-creative process. On the other hand, quality of user interaction refers to the impact of the design suggestions on the creative process. In each session, the map instance is shown sequentially based on the user's action. The patterns of the user actions indicate that they prefer a symmetric map both during and after the process. 

For the formative evaluation of the creative system, the authors reviewed the user interaction logs from the Sentient Sketchbook system. Each step of the creative process resulted in a slice of what the authors refer to as the 'creation path' that visually depicts the user's journey of creating a game level from start to finish. The authors investigate this formative data to identify different patterns and trends during the user's interaction process. 

\subsection{Evaluating creativity in 3Buddy} The evaluation of this system is done by both the users and the experts through summative user studies. Users are asked to give a value to the two metrics of evaluation called utility and efficiency. Utility refers to the ability of the system to contribute useful content. Efficiency refers to the degree in which the co-creative tool can produce useful and coherent content. 

The user study conducted to evaluate 3Buddy (both surveys and interview questions) focused on how easy the system was to use, including utilizing the various features of the tool. This type of usability analysis is interesting to evaluate the effectiveness of the tool, but it does not reveal insights about the creativity of the user or the system throughout the co-creation process. To further augment this type of investigation, the authors could employ a protocol analysis to observe the user and system behavior through time, similar to the concept of 'creation path' introduced by \cite{yannakakis2014mixed}.

\subsection{Evaluating creativity in CAHOOTS} The evaluation of this system is done by the users through summative controlled user studies. In order to test the usability of the system, participants are first introduced to the design of the system and are asked to perform a conversation for 10 minutes. Then the pairs of participants are presented with three variants of the system and are asked to chat for 10 minutes. By the end of the study, participants are required to fill out a survey in order to evaluate both the conversation and the system. The results show that participants were able to be involved in the conversation as well as finding the conversation to be funny. They also felt close to their partner as well as being able to express their sense of humor during the conversation. In order to address the qualitative analysis, the participants feedback on both prototyping and experimental phases is gathered. The results show that the feedback was positive. 

The experiments conducted to evaluate CAHOOTS focused on  usability and enjoyment, comparing it to standard text messaging applications. This type of usability analysis can reveal user satisfaction with the system, but the authors did not discuss how to evaluate the creativity of the system. Additional considerations could investigate how the suggestions of the system influence the creativity of the user and how creative the user thinks the system is in different conversational contexts. 

\subsection{Evaluating creativity in SHIMON} The evaluation of this system is done through a live performance with 160 attendants for seven minutes through a summative user studies. In this performance the robot, Shimon, with the gesture-based improvisation is shown to the audience. During the performance, a human pianist performs an opening phrase, then the robot detects the phrase and responds with preliminary gestures. This performance has three segments: The first is an open-ended collaboration between the human pianist and the robot player, Shimon. In the second phase, the robot plays in opportunistic overlay improvisation. In the last phase the robot uses a rhythmic phrase-matching improvisation. 

The authors describe a performance-based evaluation of the SHIMON system during which an audience observed the system in action as it was improvising with users. The evaluation included analyzing how the system behaved during the performance as well as audience reactions to the performance. The results of the authors analysis show that there was an alternating inspiration between the human and the robotic player. The authors also note that a video recording of the performance was widely acclaimed by the press and viewed over 40,000 times. This type of evaluation falls under the 'observation' category in our framework because the authors were working to understand how the audience perceived the performance. In the future the authors are interested to evaluate the system's gestures as well as the effect of the robotic player on band-members and audience. 

\section{Conclusions}

This paper provides a framework for evaluating creativity in computational co-creative systems. The framework provides a structure for comparing the evaluation of co-creative systems across specific examples and implementations, as well comparing to other types of creative systems, such as autonomous creative systems and creativity support tools. By asking questions such as who evaluates, when does evaluation occur, what is evaluated, and how evaluation is performed, we can broaden the scope of evaluation studies and apply methods from one area of computational creativity to another area. 

In our study of evaluation in existing co-creative systems we found a dominant focus on evaluating the user experience and the product of the experience. This demonstrates that many existing co-creative systems extend creativity support tools to include more pro-active contributions from the computational system. 

Unlike creativity support tools, co-creative systems have the potential for self-evaluation by embedding a self-awareness of the creativity of the AI agent. With a focus on evaluating the creativity of the AI agent, the computational contributions to the collaboration can be directed by its perception of the creative product. The capacity for self-evaluation can guide  users towards or away from particular regions of the space of possibilities intentionally based on the the AI agent's concept of creativity. 

Unlike autonomous creative systems, co-creative systems have the benefit of human interaction that can introduce the human perception and evaluation of the creative product during the process. Such a co-creative system requires flexibility, interruptibility, and transparency. Different strategies for achieving co-creativity include turn taking, framing, and explainable AI techniques. These strategies highlight the importance of accommodating when the AI agent has a particular intent or goal that is at odds with the user. Co-creative systems containing agents as partners will require communication of rationale and justification in order to achieve the kind of co-creativity sessions we would expect when it is among people only.  

Unlike fully autonomous creative systems and creativity support tools, the creative process used by co-creative systems is not the result of a single agent, instead it is a collaboration. This means existing approaches to evaluating computational creativity or HCI approaches to evaluate creativity support are insufficient. This identifies a new focus for research in computational creativity to study how creativity can be evaluated in human/AI collaboration with the combination and intersection of usability and creativity metrics. Evaluative methods and metrics are a step towards self-aware and intentional co-creative agents.

\bibliographystyle{iccc}
\bibliography{references}

\end{document}